\documentclass{article}

\usepackage{aisi-arxiv}
\usepackage{enumitem}

\usepackage[utf8]{inputenc}
\usepackage[T1]{fontenc}
\usepackage{charter}
\usepackage{lmodern}

\usepackage[font=small,labelfont=bf]{caption}
\usepackage{subcaption}

\DeclareCaptionLabelFormat{subfigurelabel}{Figure \thefigure(#2)}
\usepackage{textcomp}

\usepackage{hyperref}
\hypersetup{
    colorlinks=true,
    linkcolor=blue,
    citecolor=blue,
    urlcolor=blue,
}
\usepackage{subcaption}
\usepackage{url}
\usepackage{booktabs}
\usepackage{amsfonts}
\usepackage{nicefrac}
\usepackage{microtype}
\usepackage{float}
\usepackage{fancyhdr}
\usepackage{graphicx}
\usepackage{amsmath}
\usepackage{amssymb}
\usepackage{tabularx}
\newcolumntype{C}{>{\centering\arraybackslash}X}
\usepackage{longtable}
\usepackage{xcolor}
\usepackage{titlesec}
\usepackage{xspace}
\usepackage{multirow}
\usepackage[nameinlink,capitalize]{cleveref}
\usepackage{placeins}
\usepackage[round,authoryear]{natbib}

\title{Automated Alignment is Harder Than You Think}
\author{
  Aleksandr Bowkis \\
  AI Security Institute \\
  \texttt{aleksandr.bowkis@dsit.gov.uk}
  \And
  Marie Davidsen Buhl \\
  AI Security Institute \\
  \texttt{marie.buhl@dsit.gov.uk}
  \And
  Jacob Pfau \\
  AI Security Institute \\
  \texttt{jacob.pfau@dsit.gov.uk}
  \And
  Geoffrey Irving \\
  AI Security Institute \\
  \texttt{geoffrey.irving@dsit.gov.uk}
}
\begin{document}

\maketitle                                                                                                        
\let\svthefootnote\thefootnote                         
\let\thefootnote\relax
\let\thefootnote\svthefootnote 

\begin{abstract}
A leading proposal for aligning artificial superintelligence (ASI) is to use AI agents to automate an increasing fraction of alignment research as capabilities improve. We argue that, even when research agents are not scheming to deliberately sabotage alignment work, this plan could produce compelling but catastrophically misleading safety assessments resulting in the unintentional deployment of misaligned AI. This could happen because alignment research involves many hard-to-supervise fuzzy tasks (tasks without clear evaluation criteria, for which human judgement is systematically flawed). Consequently, research outputs will contain systematic, undetected errors, and even correct research could be incorrectly aggregated into overconfident safety assessments. This problem is likely to be worse for automated alignment research than for human-generated alignment research for several reasons: 1) optimisation pressure means agent-generated mistakes are concentrated among those that human reviewers are least likely to catch; 2) agents are likely to produce errors that do not resemble human mistakes; 3) AI-generated alignment solutions may involve arguments humans cannot evaluate; and 4) shared weights, data and training processes may make AI outputs more correlated than human equivalents. Therefore, agents must be trained to reliably perform hard-to-supervise fuzzy tasks. Generalisation and scalable oversight are the leading candidates for achieving this but both face novel challenges in the context of automated alignment.
\end{abstract}


\hypertarget{introduction}{%
\section{Introduction}\label{introduction}}

Alignment of artificial superintelligence (ASI) is an important, unsolved research problem\footnote{The definition of what constitutes ASI, and the degree to which misaligned ASI poses a significant risk, is contested.}. AI research agents may help solve this problem, for example via the following plan:

\begin{itemize}
\item
  Build AI agents that can do empirical alignment work (e.g.~writing code, running experiments, designing evaluations and red teaming) and confirm they are not scheming\footnote{We consider an agent to be scheming if it covertly and strategically pursues a hidden objective \citep{meinke2025frontiermodelscapableincontext}. This definition permits isolated misaligned behaviours such as reward hacking. Unlike other common definitions (e.g., \citet{carlsmith2023scheming}), our definition is not restricted to behaving well in training in order to gain power later.}.
\item
  Use these agents to build increasingly sophisticated empirical safety cases for each successive generation of agents, gradually automating more of the research process.
\item
  Hand over primary research responsibility once agents outperform humans at all relevant alignment tasks.
\end{itemize}

\begin{figure}[H]
\centering
\begin{subfigure}{0.85\linewidth}
    \centering
    \includegraphics[width=\linewidth]{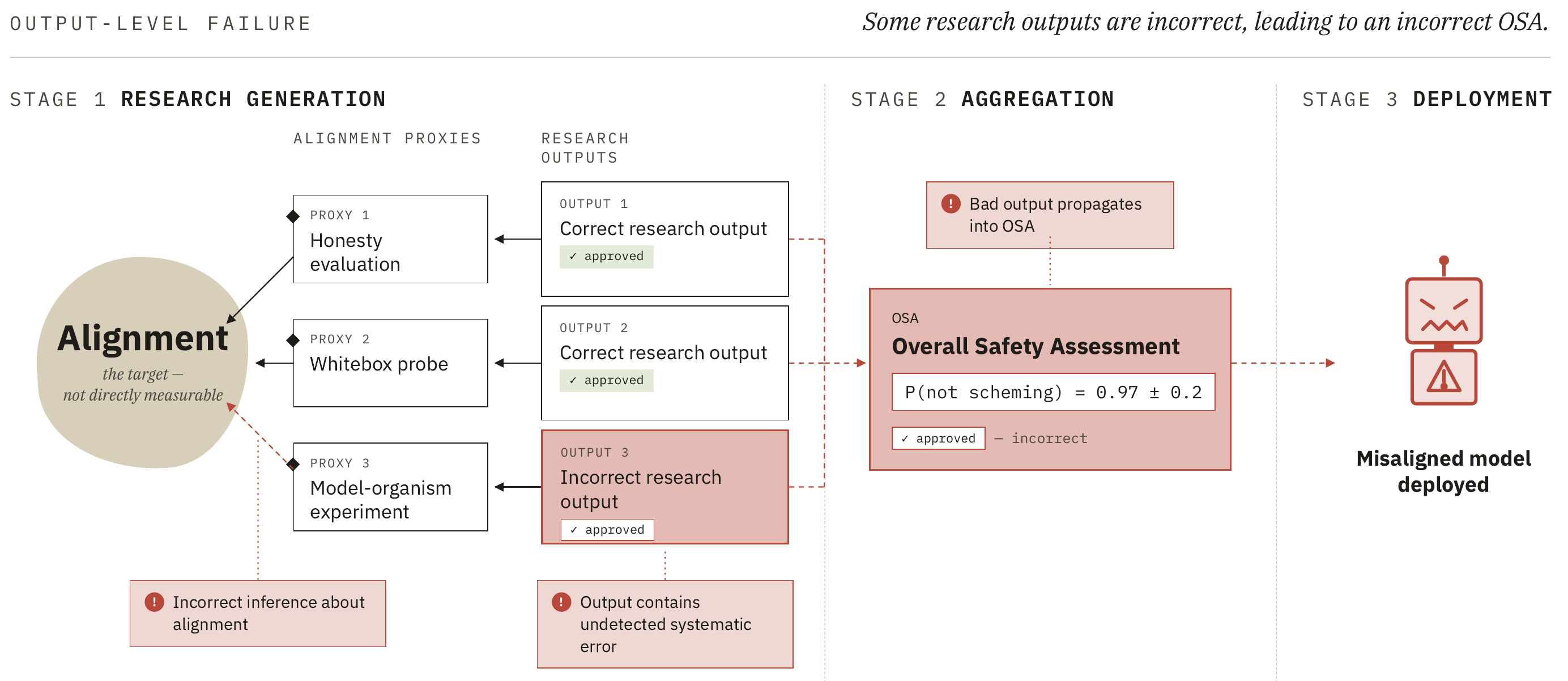}
    \caption{Output-level failures: undetected systematic errors in individual research outputs.}
    \label{fig:1a}
\end{subfigure}

\vspace{0.5em}

\begin{subfigure}{0.85\linewidth}
    \centering
    \includegraphics[width=\linewidth]{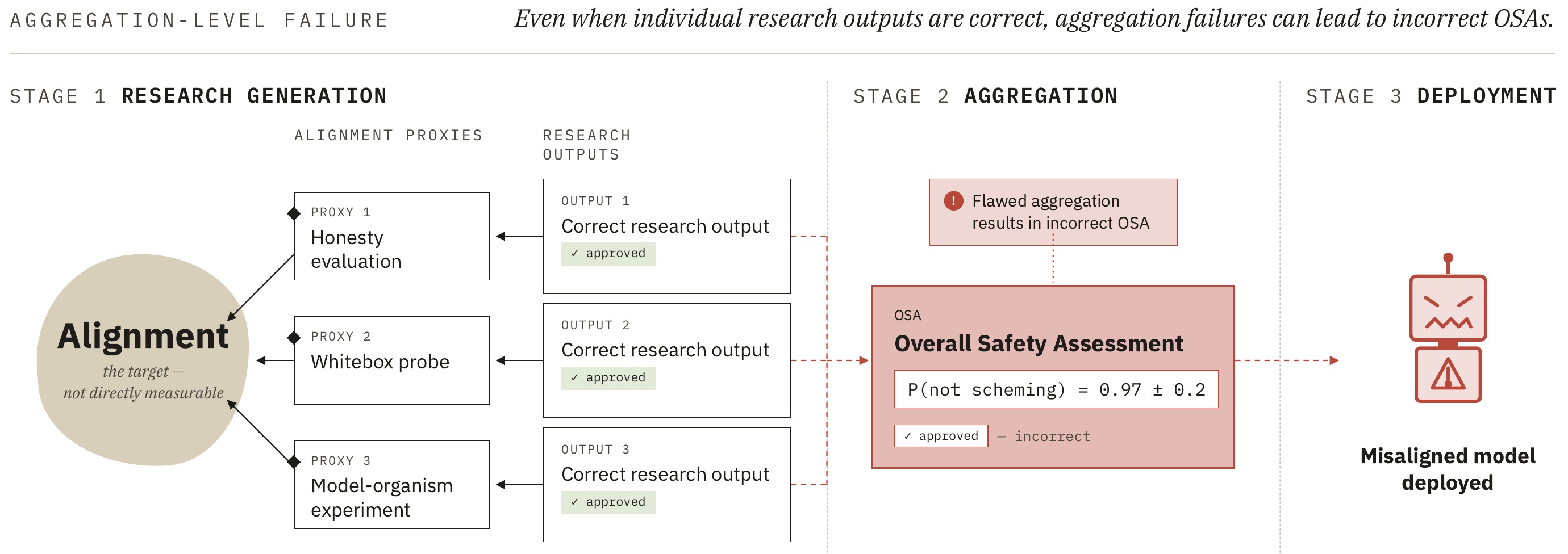}
    \caption{Aggregation-level failures: flawed combination of correct research outputs.}
    \label{fig:1b}
\end{subfigure}

\caption{Automated alignment can fail, leading to the deployment of a misaligned AI, if research outputs contain undetected systematic errors (\subref{fig:1a}) or if the aggregation of research outputs is flawed (\subref{fig:1b}). Since alignment cannot be measured directly, research instead targets alignment proxies (model organisms, honesty evaluations, whitebox probes etc.). Incorrect judgement about the implications of results for alignment can lead to systematic, undetected errors in individual research outputs. Aggregation of outputs can lead to incorrect overall safety assessments, even when individual outputs are correct, if the correlation between the uncertainties in each research output is mis-modelled (Sec.~\ref{aggregation-level-failures}).}
\label{fig:1}
\end{figure}

This paper argues that automating alignment research in this manner could produce catastrophically misleading safety cases, causing researchers to believe that an egregiously misaligned AI is safe, even if AI agents are not scheming to deliberately sabotage alignment research. Our core argument (Fig. \ref{fig:1}) is as follows:

\begin{enumerate}
\def\labelenumi{\arabic{enumi}.}
\item
  The goal of an automated alignment program is to produce an \textbf{overall safety assessment (OSA)}\footnote{"Overall safety assessment" is intended as a more general term than "safety case"; a safety case is an OSA, but an OSA could also look like a risk report, a compilation of empirical evidence, or a judgement by an individual human or AI).} - an estimate of the probability that the next-generation agent is non-scheming\footnote{If the next-generation agent will also be used to do automated alignment research, the OSA must additionally estimate the probability that its research outputs can be assessed well enough to produce a calibrated OSA for the next-next-generation.} - that is both \textbf{calibrated} and \textbf{shows low risk}.
\item
  Producing an OSA involves several tasks that are difficult to check. We refer to these as \textbf{hard-to-supervise fuzzy tasks:} \emph{tasks without clear evaluation criteria, for which human judgement is systematically flawed}. Two particularly important such tasks are\footnote{Another important, hard-to-supervise fuzzy task is \textbf{steering a body of research} and allocating resources efficiently between research directions. This involves assessing whether individual research directions are worth pursuing and whether the research corpus is sufficiently broad.}:

  \begin{enumerate}
  \def\labelenumii{\arabic{enumii}.}
  \item
    \textbf{Measuring what you care about: }The alignment properties of a sufficiently advanced AI are not directly measurable. We cannot safely deploy the AI in situations where it could cause serious harm and observe whether it takes misaligned actions. Instead, research must target alignment proxies (e.g.~honesty evaluations, model organism experiments, whitebox probes) and then make a judgement about how relevant the results are for alignment of the AI.
  \item
    \textbf{Aggregating correlated evidence: }OSAs will likely rely on multiple pieces of evidence. If each piece of evidence is independent, aggregating them into an OSA would be straightforward. In practice, however, there will be uncertainty about the conclusions from each research output and these uncertainties will be correlated. This happens because research outputs share assumptions and methodologies; because research agents share weights, training processes and data; and because human researchers share biases. If these correlations are mis-modelled, the resulting OSA could be catastrophically flawed even if no individual piece of evidence is wrong.
  \end{enumerate}
\end{enumerate}

Because these tasks are hard to supervise, research outputs (including OSAs) will contain \textbf{systematic, undetected errors}, regardless of whether these tasks are completed by human researchers or research agents trained using human data. We already see examples of such errors in current models, especially on hard tasks.

\begin{enumerate}
\def\labelenumi{\arabic{enumi}.}
\setcounter{enumi}{2}
\item
  Of course, such errors are also present in human-generated alignment research. However, \textbf{errors in automated alignment research are likely to be harder to identify than the human baseline}. There are five reasons for this:

  \begin{enumerate}
  \def\labelenumii{\arabic{enumii}.}
  \item
    \textbf{Optimisation pressure: }AI research is optimised for human approval to a greater extent than human research, and so any errors are concentrated among those that human reviewers are least likely to catch.
  \item
    \textbf{Alien mistakes: }Agents may introduce errors that do not resemble human mistakes. Reviewers may struggle to identify these, even if they are in principle easy to correct.
  \item
    \textbf{More correlated research: }Shared weights, training processes and data sources for agents introduce additional correlation structure relative to human-generated research.
  \item
    \textbf{Research volume: }OSAs based on AI research may reference much larger evidence bases, increasing the potential for complex and hard-to-spot correlations between research outputs.
  \item
    \textbf{Non-human-evaluable arguments: }Alignment solutions for ASI may rely on arguments that humans are unable to follow, making errors impossible to identify.
  \end{enumerate}
\end{enumerate}

In most domains, iteration is able to correct for undetected errors. Mistakes not caught during one experiment are often surfaced by subsequent research or real-world system behaviour. Unfortunately, alignment lacks the safe feedback loops that are required for such an error-correction process to work: producing an overly optimistic OSA could result in the deployment of a misaligned AI before the error is caught, which could be catastrophic. Therefore, we need to train agents to reliably perform well on these hard-to-supervise fuzzy tasks the first time. Direct training using human feedback on these hard-to-supervise fuzzy tasks is not sufficient as human approval does not indicate correctness. Two alternatives remain:

\textbf{Generalisation: }Train agents on easier-to-supervise training proxies and rely on generalisation to the hard-to-supervise fuzzy research tasks we care about. This requires us to predict generalisation behaviour, since performance on the hard-to-supervise fuzzy task cannot be evaluated directly.

\textbf{Scalable oversight: }Improve the reward signal for hard-to-supervise fuzzy tasks, for example by decomposing them into easier-to-supervise subtasks. Existing protocols (e.g.~recursive reward modelling or debate) may not work, because they do not have a good solution to the problem of aggregating correlated evidence.

The paper proceeds as follows. \cref{an-automated-alignment-research-program} sketches an automated alignment program as background context for the rest of the paper. \cref{the-undetected-errors-failure-mode} explains why hard-to-supervise fuzzy tasks are pervasive in alignment research and why this leads to more undetected, systematic errors than the human baseline. \cref{undetected-errors-could-be-hard-to-prevent} argues that these errors will be hard to prevent and outlines the open research problems in scalable oversight and generalisation. \cref{conclusion} concludes with future research directions.

\hypertarget{an-automated-alignment-research-program}{%
\section{An automated alignment research program}\label{an-automated-alignment-research-program}}

To ground our analysis, this section paints a rough picture of what an automated alignment research program might look like. The plan is inspired by (but not identical to) plans proposed elsewhere such as \cite{carlsmith2025can, clymer2025how, leike2022minimal}.

We model AI development as a discrete sequence of agents, indexed by iteration \emph{N}, with capabilities increasing monotonically. Each agent is released with a fixed capability profile and is eventually superseded by its successor. At each iteration, an \textbf{automated alignment research program} uses multiple instances of Agent \emph{N}, together with human researchers, to research alignment techniques for the next-generation model, Agent \emph{N+1}. The final output of the research process is an \textbf{overall safety assessment (OSA)}: a calibrated estimate of the probability \(P(\text{Agent } N{+}1 \text{ is not scheming} \mid \text{evidence})\). If the OSA is good enough, Agent \emph{N+1} is deployed---including as part of the workforce that produces the OSA for Agent \emph{N+2}. As \emph{N} grows, humans delegate a greater fraction of the research process to agents.

This plan requires that the initial Agent 0 is not scheming. The plan then \emph{bootstraps }its way towards an OSA for an arbitrarily powerful ASI: Agent 0 provides the evidence that Agent 1 is not scheming, Agent 1 provides the evidence that Agent 2 is not scheming, and so on. The hope is that if the capability gap remains small at each stage the additional research required to align Agent \emph{N+1} is within the scope of Agent \emph{N}.

\begin{figure}[h]
\centering
\includegraphics[width=0.85\linewidth]{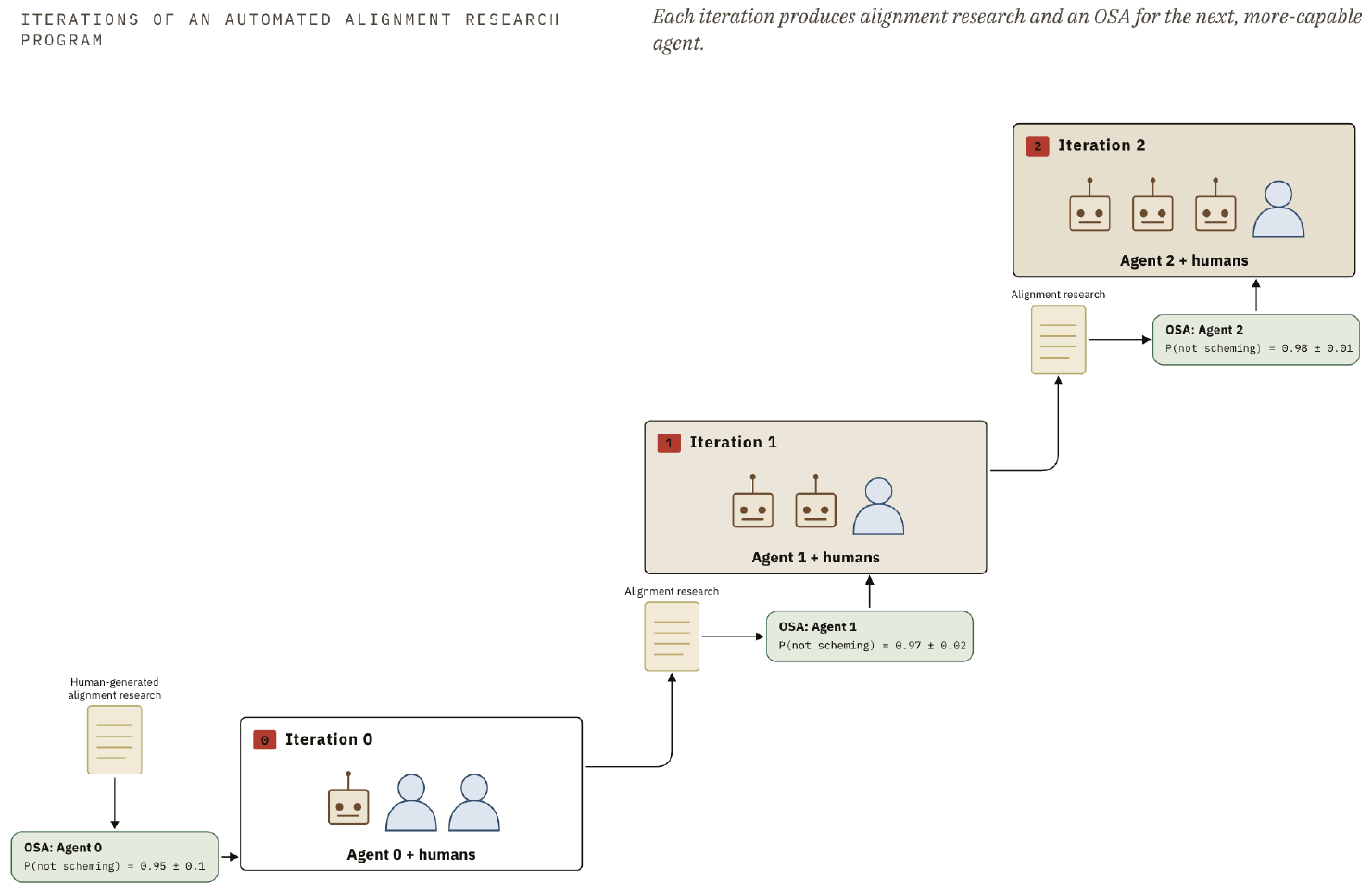}
\caption{An automated alignment research program (AARP) uses both human and Agent research labour to produce alignment research with the goal of combining this into an overall safety assessment (OSA): a calibrated probability estimate that the next generation agent is not scheming. As automation progresses, agents perform a greater fraction of research tasks.}
\label{fig:2}
\end{figure}

\hypertarget{the-undetected-errors-failure-mode}{%
\section{The `undetected errors' failure mode}\label{the-undetected-errors-failure-mode}}

Two of the ways in which the above plan could fail are:

\begin{enumerate}
\def\labelenumi{\arabic{enumi}.}
\item
  \textbf{Research sabotage: }If the initial OSA is incorrect and Agent 0 is scheming, it could undermine alignment research, resulting in further alignment failures.
\item
  \textbf{No alignment solution is found: }The combined efforts of Agent N and human researchers may be insufficient to find an alignment solution for Agent N+1. In this case we fail to align Agent N+1 but are aware of our failure.
\end{enumerate}

While these failure modes are important, we set them aside for the remainder of the paper. We argue that \emph{even absent these problems}, there is a third reason automated alignment research might fail:

\def\labelenumi{\arabic{enumi}.}
\begin{quote}
    \textbf{Undetected systematic errors: }At any stage, the research outputs and the OSA produced by the automated alignment program could appear compelling but in fact contain critical, undetected errors. In this case, we fail to align Agent N+1 but are (incorrectly) confident that we have succeeded.
\end{quote}

This section presents the case for worrying about undetected systematic errors. Our core claims are that alignment research involves important fuzzy tasks (Sec. \ref{alignment-research-involves-important-fuzzy-tasks}), that many of these tasks are hard-to-supervise (Sec. \ref{alignment-involves-many-hard-to-supervise-fuzzy-tasks}) and that this causes automated alignment research to contain systematic errors that both humans and agents fail to spot, leading to false confidence in alignment (Sec. \ref{hard-to-supervise-fuzzy-tasks-lead-to-undetected-errors}).

\hypertarget{alignment-research-involves-important-fuzzy-tasks}{%
\subsection{Alignment research involves important fuzzy tasks}\label{alignment-research-involves-important-fuzzy-tasks}}

Following \cite{leike2024crisp}, we can think of research tasks as lying on a spectrum from crisp to fuzzy.

\textbf{Crisp tasks }have clear evaluation criteria: success is verifiable and experts reliably agree on correct answers --- for example, proving a mathematical result or writing code for an experiment.

\textbf{Fuzzy tasks} lack clear evaluation criteria --- whether an output is good is harder to specify, and reasonable experts may disagree. Examples include identifying objects in images or reaching a conclusion about the implications of a research paper.

Fuzzy tasks are particularly prevalent in alignment research compared to capabilities research. This is because for capabilities research, the properties that are being targeted are generally easy to measure, resulting in tight feedback loops: a model is trained, its performance is evaluated against a benchmark, the training process is updated and this process repeats. In alignment, the true feedback signal (whether a deployed model takes a misaligned action) is not available. Research must therefore target alignment proxies: behavioral evaluations, whitebox signals, red teaming results, testing techniques on weaker, safer models etc. Inferring conclusions about alignment from these proxies is a difficult fuzzy judgement.

AI agents will likely perform increasingly many and difficult fuzzy tasks as automation progresses. For illustration, consider the following two sketches of how fuzziness might enter during the early and late stages of automated alignment research.

\hypertarget{early-stages-of-automation}{%
\subsubsection{Early Stages of Automation}\label{early-stages-of-automation}}

An early stage of automation might look like this:

\textbf{AI agents} perform relatively open-ended research tasks, for example optimising against alignment metrics \citep{wen2026automated}, generating evaluation environments \citep{petri2025}, and monitoring model outputs for misaligned behaviour \citep{gasteiger2025sandbag, greenblatt2023ai}.

\textbf{Humans }set the targets and high-level research directions. They also assemble OSAs from AI-generated and human-generated evidence (this could even be implicitly via surveys of research staff \citep{anthropic2026opus46card}). An early-stage OSA might consist of a large body of experiments showing a model behaves in an aligned manner across diverse contexts, supplemented by whitebox evidence about internal representations.

At this stage, humans perform most of the load-bearing fuzzy tasks: judging how AI-generated research generalises beyond the experimental settings, assessing whether models are Goodharting metrics \citep{manheim2018goodhart} rather than improving on the underlying alignment properties, evaluating how representative the experimental coverage is, and making an overall judgement about whether the evidence as a whole is sufficient. While agents perform tasks that appear crisp, those tasks contain many judgement calls about details. For example, automated red teaming involves calls such as which behaviours to look for, how to construct an environment, how to interpret a given trajectory, and how to present results. These calls impact bigger fuzzy tasks, such as judging experiment validity and correlations.

\hypertarget{later-stage-of-automation}{%
\subsubsection{Later Stage of Automation}\label{later-stage-of-automation}}

A later stage of automated alignment research might look like this:

\textbf{AI agents} generate new research paradigms and metrics (for example, generating ideas such as weak-to-strong generalisation and metrics such as performance gap recovered \citep{burns2023weak}. They also propose directions within those paradigms and increasingly assemble OSAs themselves. These OSAs may initially look similar to human OSAs but increasingly drift towards using concepts and structures that are unfamiliar to humans.

\textbf{Humans} evaluate AI-generated research outputs, including OSAs, provide high-level steering and decide resource allocation. The role of human researchers at this stage is analogous to a funding body evaluating the outputs from a research grant.

Agents now perform many of the load-bearing fuzzy research tasks.

\hypertarget{alignment-involves-many-hard-to-supervise-fuzzy-tasks}{%
\subsection{Alignment involves many hard-to-supervise fuzzy tasks}\label{alignment-involves-many-hard-to-supervise-fuzzy-tasks}}

\begin{figure}[H]
\centering
\includegraphics[width=0.95\linewidth]{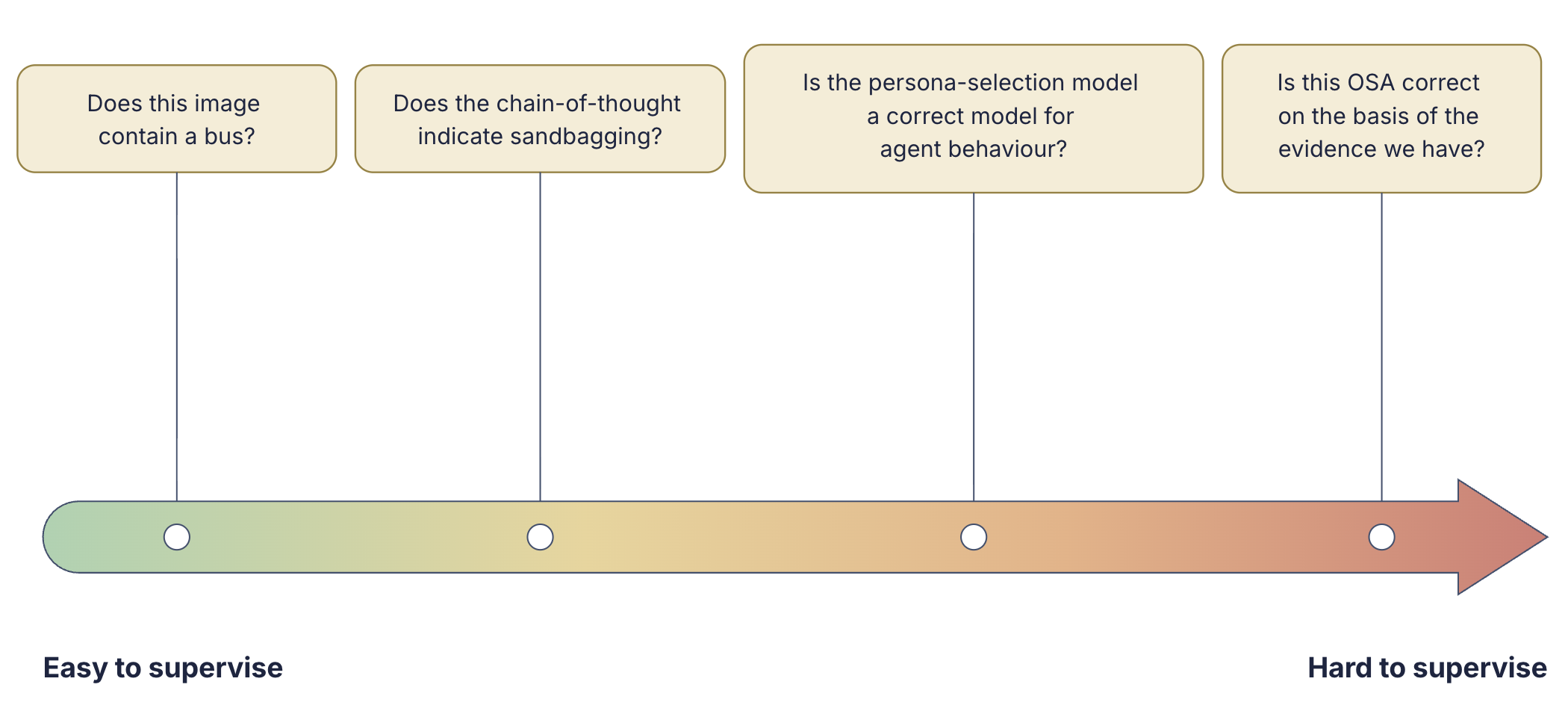}
\caption{Examples of tasks placed on the easy-to-supervise vs hard-to-supervise spectrum.}
\label{fig:spectrum}
\end{figure}

A key property of fuzzy tasks is how difficult they are for humans to supervise. Some fuzzy tasks are easy for humans to judge, even if it is hard to specify precisely why an output is correct (e.g.~identifying whether an image contains a bus). Other fuzzy tasks are \textbf{hard to supervise}: human judgement about whether an output is correct is systematically flawed (Fig. \ref{fig:spectrum})    .

Many research tasks are examples of hard-to-supervise fuzzy tasks. We can see this from the history of science where human judgement has been consistently unreliable: whole fields have refused to adopt well-supported ideas, or expended huge effort investigating unproductive hypotheses.

For example, the existence of a luminiferous aether (a medium through which light waves were posited to travel) was widely accepted in 19th century physics despite the lack of supporting evidence. It was only the experiments of Michelson and Morley in 1887 and Einstein's theory of special relativity in 1905 that forced the scientific community to abandon these ideas.

Despite many such historical examples of poor individual judgement, science has generally progressed over time. Physics, for example, has produced quantum mechanics, general relativity, the standard model of particle physics and a huge number of associated technological and scientific breakthroughs. This is because iteration has often corrected errors of judgement: multiple hypotheses were made, built upon, invalidated and then revised.

The effectiveness of this error-correction mechanism varies by domain: it works well in fields where hypotheses are easy to test, such as physics. However, in fields where hypotheses are hard to verify with high confidence, such as psychology, it is less effective \citep{osc2015estimating, kahneman2009conditions}. Furthermore, error-correction through iteration does not ensure that a field continually progresses; the overall credibility of a field can fluctuate over time \citep{ioannidis2012why}. This is also true for alignment: \textbf{At any given moment the body of alignment research could contain mostly correct or mostly incorrect research and we may be unable to distinguish between the two}.

Unfortunately, alignment lacks the safe feedback loops that are required for error-correction to work \citep{griffin2026unsafetoverify}: Producing an overly optimistic OSA could result in the deployment of a misaligned AI before the error is caught. This means alignment cannot rely on iteration to correct errors of judgement and we therefore need reliable methods for training agents to perform well on hard-to-supervise fuzzy tasks. Despite considerable effort, there has been minimal progress on this so far \citep{leike2025solvable}.

Alignment research tasks are also likely to become even harder to supervise over time. As agents become more capable they are able to produce more complex outputs (for example, by designing more elaborate experiments) that become increasingly difficult to evaluate. The space of possible misaligned strategies also grows as capabilities improve. This increases both the number of alignment properties that must be tested for and the difficulty of identifying and preventing them. For example, a sufficiently advanced misaligned model may pursue strategies such as gradient hacking \citep{hubinger2019gradient} and these strategies could be harder to detect.

\hypertarget{hard-to-supervise-fuzzy-tasks-lead-to-undetected-errors}{%
\subsection{Hard-to-supervise fuzzy tasks lead to undetected errors}\label{hard-to-supervise-fuzzy-tasks-lead-to-undetected-errors}}

We have argued that the formation of OSAs involves hard-to-supervise fuzzy tasks. Below, we explain why this can lead to errors in OSAs, regardless of whether those tasks are performed by humans or AI agents trained on human supervision. These errors can occur via two mechanisms:

\textbf{Output-level failures:} Agents produce incorrect research outputs when completing hard-to-supervise fuzzy tasks.

\textbf{Aggregation-level failures:} Even when individual research outputs are correct up to their stated uncertainty, these uncertainties will be correlated. Mis-modelling this correlation structure when aggregating research outputs into OSAs leads to incorrect conclusions.

\hypertarget{output-level-errors}{%
\subsubsection{Output-level failures}\label{output-level-errors}}

If an AI agent is trained on a systematically flawed human supervision signal for some research tasks, it will make mistakes when performing those research tasks. As a result, AI-generated research is likely to contain systematic errors - and specifically errors that subsequent human review is unlikely to catch. 

Current frontier model outputs already make many errors on complex research output. Examples include compelling but false mathematical proofs \citep{feldman2025godel}, overselling incomplete work \citep{greenblatt2026misaligned}, and reward hacking \citep{metr2025rewardhacking}. These behaviors appear to get worse on harder tasks --- in the limit, on impossible coding tasks, Claude Opus 4.7 attempts to cheat tests 45\% of the time by default \citep{anthropic2026opus47} and GPT-5.5 lies about having completed the task in 29\% of samples \citep{openai2026gpt55}. Many such errors are currently relatively easy to catch, but errors could become harder to catch in future, as AI agents are tasked with harder-to-verify tasks.

Of course, human research also contains errors. However, there are reasons to think that errors in AI-generated research will be systematically harder to spot:

\begin{itemize}
\item
  \textbf{Alien mistakes: }Agent behaviour differs from human behaviour on equivalent tasks because their cognitive architecture is fundamentally different. As a result, they are likely to produce errors that do not resemble human errors, which reviewers may struggle to identify. For example, a research agent might name a scheming evaluation scheming\_test.py, leaking details of the evaluation to the model being tested, something a human researcher is unlikely to do. Less obviously, since models can detect authorship from subtle clues (a phenomenon sometimes referred to as ``true sight'' \citep{panickssery2024llm}) this kind of leakage can occur through channels a human reviewer would not check.
\item
  \textbf{Optimisation pressure:} Training on human feedback optimises AI-generated research to appear correct to human judges. On hard-to-supervise fuzzy tasks, human feedback does not necessarily indicate correctness. Any errors will resemble those that humans do not spot. Optimisation pressure could also make errors harder and harder to spot over time.
\item
  \textbf{Non-human-evaluable arguments:} Alignment solutions for ASI may rely on arguments that humans are unable to follow. For example, the arguments could rely on new concepts or ontologies that are alien from human concepts or ontologies; or they could rely on alien heuristics learned by recognising patterns across vasts amount of data.
\end{itemize}

As a result, it could be the case that \textbf{AI-generated alignment research (that has been through human review) contains more errors than equivalent human research}. 

An additional challenge is that AI agents can generate larger volumes of research more quickly. This means that there are more total errors - but also that human reviewers may face pressure to quickly review large quantities of AI-generated work, making them even less likely to catch mistakes \citep{goddard2012automation}.

Agent errors are also amplified when agents begin to both define and optimise against alignment metrics. Currently human researchers generate alignment metrics (for example the ``performance gap recovered (PGR)'' used in Anthropic's weak-to-strong research \citep{burns2023weak}) and agents optimise against them \citep{wen2026automated}. At some point, however, research agents will produce their own alignment metrics. Like other AI-generated outputs, these metrics are more likely to contain hard-to-spot flaws than human-generated alternatives. These flaws are then propagated across all subsequent research that optimises against the agent-proposed metric.

\hypertarget{aggregation-level-failures}{%
\subsubsection{Aggregation-level failures}\label{aggregation-level-failures}}

There will always exist uncertainty about how correct a given research paper is. This uncertainty is necessarily correlated between different papers\footnote{This problem of correlated uncertainty was discussed in learning the prior \citep{christiano2020learningprior} and learning the smooth prior \citep{irving2022smoothprior}.}. Mis-modelling this correlation structure can lead to undetected errors when aggregating research outputs into OSAs and subsequent alignment failures, even when individual research outputs are correct.

For example, imagine that Agent N has produced ten research papers, each providing evidence that Agent N+1 is not scheming. Treated as independent evidence these papers indicate that the likelihood that Agent N+1 is scheming is low. If, however, these papers share an unknown correlation (perhaps they all assume that the persona selection model \citep{anthropic2026psm} is the correct framework for understanding agent behaviour) the true probability of scheming is much higher. Failing to account for this would result in overconfident OSAs.

There are three sources of correlated uncertainty:

\begin{itemize}
\item
  \textbf{Agents: }Shared weights, architectures, training data, and optimisation pressure from the training process (for example, agents trained against the same reward model inherit its systematic biases).
\item
  \textbf{Human reviewers: }Shared systematic biases across human reviewers.
\item
  \textbf{The research process: }Shared assumptions, methodologies and conceptual frameworks.
\end{itemize}

Some of these correlations are easy to identify (e.g. two papers that use the same interpretability technique) but others will be unknown. The correlation structure can be complex: dependencies between groups of research papers can exist even when there are no pairwise correlations. The correlation structure can also change suddenly. For example, if Agent N+1 learns to gradient hack, papers whose conclusions relied on this being out of scope now fail together, and an aggregation process that treats them as independent will produce overconfident OSAs.

The severity of this problem depends on the volume of evidence that must be combined to form an OSA. There are two extremes: at one end a single experiment clearly proves or disproves that Agent N+1 is not scheming; at the other, all pieces of evidence must be considered holistically. The fewer pieces of evidence required, the simpler the correlation structure and the easier the aggregation process. One might therefore attempt to sidestep the problem of correlated uncertainty by using an agent to select a small sample of key research papers to be combined into an OSA. Unfortunately, this trades off the difficulty in aggregating many correlated outputs for trust in the agent's judgement about which papers are most important.

This problem is not unique to alignment research. Both forecasting and safety case generation require aggregating correlated judgements. Techniques from these fields may be helpful in solving aggregation-level errors in alignment research. For example, state-of-the-art safety case methodology proposed using Bayesian Belief Networks to map correlation structures in evidence \citep{bloomfield2020assurance}.

\hypertarget{undetected-errors-could-be-hard-to-prevent}{%
\section{Undetected errors could be hard to prevent}\label{undetected-errors-could-be-hard-to-prevent}}

There are two approaches to training an agent to perform hard-to-supervise fuzzy tasks:

\begin{itemize}
\item
  \textbf{Trust generalisation from an easier-to-supervise training proxy: }There may be easy-to-supervise tasks where performance reliably transfers to hard-to-supervise fuzzy research tasks. For example, training models to predict experimental results in alignment might generalise to predicting if the model will scheme during deployment.
\item
  \textbf{Use a scalable oversight protocol: }It may be possible to decompose hard-to-supervise fuzzy tasks into easier-to-supervise subtasks. For example, AI debate might make it possible to identify if an output is flawed by evaluating a single load-bearing node in a debate between two agents.
\end{itemize}

We discuss these approaches in sections \ref{generalisation} and \ref{scalable-oversight} respectively.

\hypertarget{generalisation}{%
\subsection{Generalisation}\label{generalisation}}

One approach for training agents to perform hard-to-supervise fuzzy tasks is to train on related tasks that are easier to supervise and hope that performance on those tasks generalises. If generalisation behaviour is predictable, this removes the need to evaluate agent performance on the hard-to-supervise tasks.

It is sometimes suggested that continued RL training on a broadening set of (largely verifiable) tasks will yield generalisation to other domains, including hard-to-supervise fuzzy tasks \citep{amodei2026dwarkesh}. Indeed, models already perform well on some fuzzy tasks, such as writing. In reality, writing is an example of an easy-to-supervise fuzzy task where humans can provide a reliable training signal. Agent performance on such tasks is therefore better explained by RL training on tasks with similar distributions using human feedback or AI self-critique. Since human feedback is an unreliable training signal for hard-to-supervise fuzzy tasks, it is unlikely that this method will work for such tasks.

Instead, one might try to generalise from easier-to-supervise training proxies. We define a true task and training proxy pair as follows:

\textbf{True task:} The task that the agent is intended to perform. Example: Design an experiment that reveals whether a model is behaving honestly.

\textbf{Training proxy: }A related task that is easier to train for than the true task. Example: Predict outcomes of existing honesty evaluations.

One might expect performance on the training proxy to generalise to the true task. In the example above, because the two tasks appear similar (both involve predicting similar model behaviours) and because in learning the training proxy, the model may have developed a detailed understanding of what drives model behaviour.

However, the training proxy only provides information about how a model behaves in the setting of the evaluation. If your evaluations do not reliably measure honesty, training may not be effective. In other words, the training proxy is no better than your alignment evaluations. The broader epistemic problem is that you cannot directly measure generalisation if you cannot evaluate performance on the true task.

There is an additional generalisation challenge beyond performance on the training proxy reliably transferring to performance on the true task. Performance on both tasks must also be maintained across the distribution shift from training to deployment\footnote{Strictly, the case that training on a proxy task generalises rests on three claims: 1. The training proxy and true tasks are tightly coupled (i.e. performance on the training proxy reliably indicates performance on the true task); 2. The training proxy and true tasks remain tightly coupled across the distribution shift from training to deployment; and 3. The performance (on either task) generalises across the distribution shift from training to deployment. We group claims 2 and 3, as both are concerned with robustness to distribution shift.}. However, we consider this less concerning as there are promising techniques for maintaining performance across distribution shifts, for example online learning (which reduces the training to deployment gap by continuing to train on deployment data).

There are two approaches to solving the problem of validating training proxies:

\begin{enumerate}
\def\labelenumi{\arabic{enumi}.}
\item
  Develop a detailed theoretical and empirical understanding of how generalisation works, in order to reliably predict generalisation from training proxies, removing the need to evaluate performance on the true hard-to-supervise task. \citep{carlsmith2025can}
\item
  Improve our ability to evaluate performance on the true hard-to-supervise task, thereby making it possible to validate the training proxy. One approach is scalable oversight, discussed in the next section.
\end{enumerate}

\hypertarget{scalable-oversight}{%
\subsection{Scalable oversight}\label{scalable-oversight}}

Another approach for training agents to perform hard-to-supervise fuzzy tasks is to directly improve the training signal on these tasks. This can be done by  directly improving the quality of human judgement. For example, by carefully selecting and training judges, designing interfaces that make review easier (e.g. by requiring agents to point to important decision points) or adopting processes that use structured deliberation or adversarial review such as the Delphi method \citep{dalkey1962delphi}. However, these are likely to be insufficient,  as human judgement is fundamentally limited. Instead, more sophisticated scalable oversight protocols may be required.

For example, a scalable oversight protocol could be used to decompose hard-to-supervise fuzzy tasks into subtasks which are easier to evaluate. Humans could then check the easier subtasks and these judgments could be propagated up to the main task. Such a scalable oversight protocol could be used to directly train an agent to perform a hard-to-supervise fuzzy task, or to determine whether an agent has generalised successfully from a training proxy\footnote{A training proxy is not strictly necessary if a scalable oversight protocol exists. However, in practice it may be more efficient to use a training proxy and verify that performance generalises using the scalable oversight protocol than to directly train on the signal from the scalable oversight protocol.}.

There are many scalable oversight proposals, for example recursive reward modelling \citep{leike2018scalable}, debate \citep{irving2018ai}, and iterated amplification and distillation \citep{christiano2018supervising}. Unfortunately, it is unclear if these proposals work for the type of hard-to-supervise fuzzy research tasks we are concerned with. There are two reasons for this: (1) subtasks will still require judgement; and (2) existing protocols are unable to recombine correlated, probabilistic subtasks.

\hypertarget{judgement}{%
\subsubsection{Judgement}\label{judgement}}

Most existing scalable oversight protocols involve decomposing hard-to-supervise fuzzy tasks into subtasks. However, some of these subtasks will themselves be fuzzy and therefore require judgement. For example, one subtask of forming an OSA might be: \emph{How relevant is experiment E (that uses model organism M) for determining if Agent N+1 is scheming? }A human judge has to decide how important evidence from the experiment is. Misjudging the importance of the experiment could invalidate any OSA that cites it as evidence.

\begin{figure}[h]
\centering
\begin{subfigure}[t]{0.48\linewidth}
  \centering
  \includegraphics[width=\linewidth]{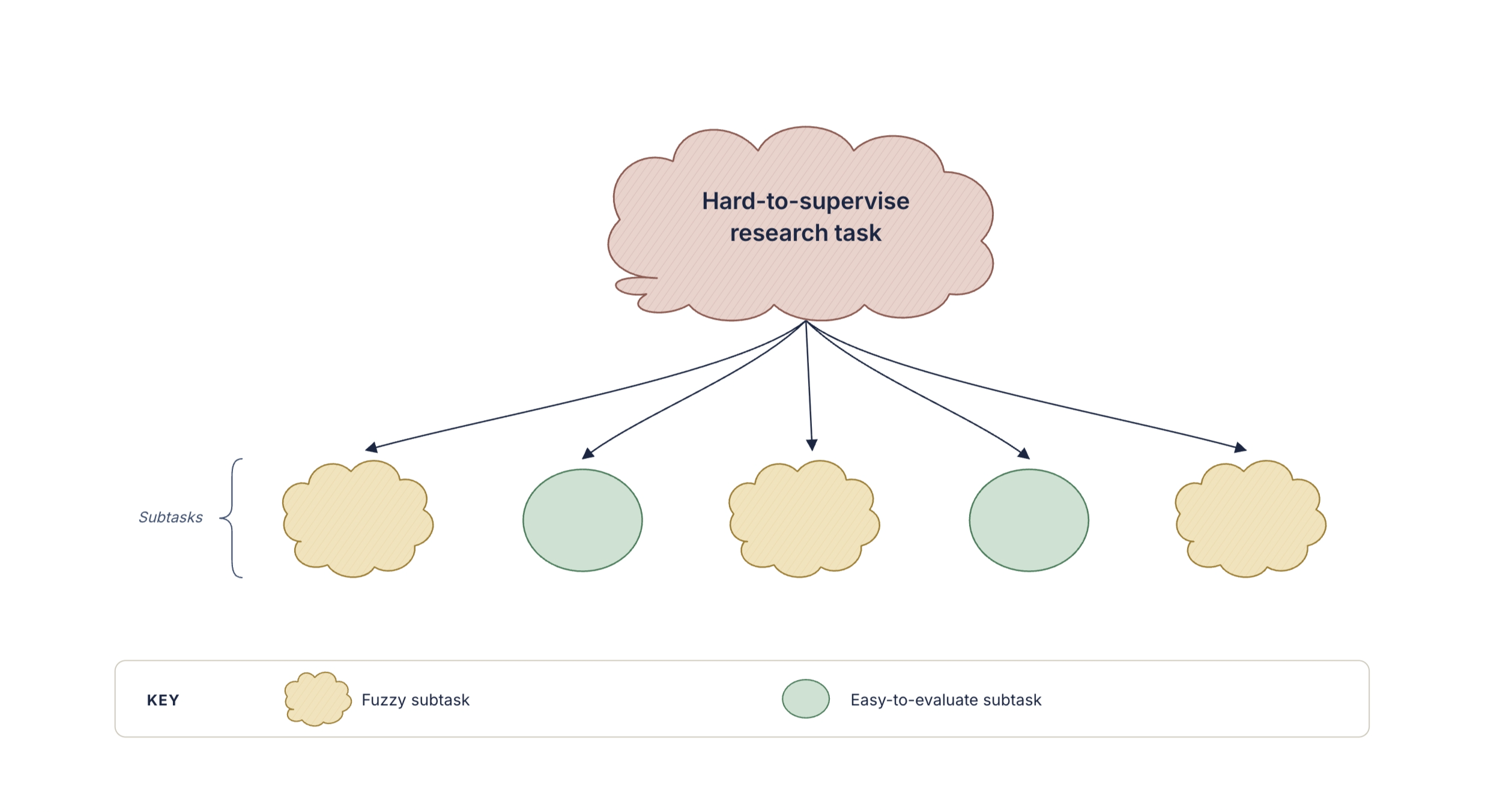}
  \caption{A scalable oversight protocol decomposes a hard-to-supervise fuzzy research task into easier subtasks. Some of these subtasks will themselves be fuzzy tasks (yellow clouds) and some will be non-fuzzy tasks that are easy to evaluate (green circles). If the fuzzy subtasks are easier to evaluate than the original task then the scalable oversight protocol is useful.}
  \label{fig:3}
\end{subfigure}\hfill
\begin{subfigure}[t]{0.48\linewidth}
  \centering
  \includegraphics[width=\linewidth]{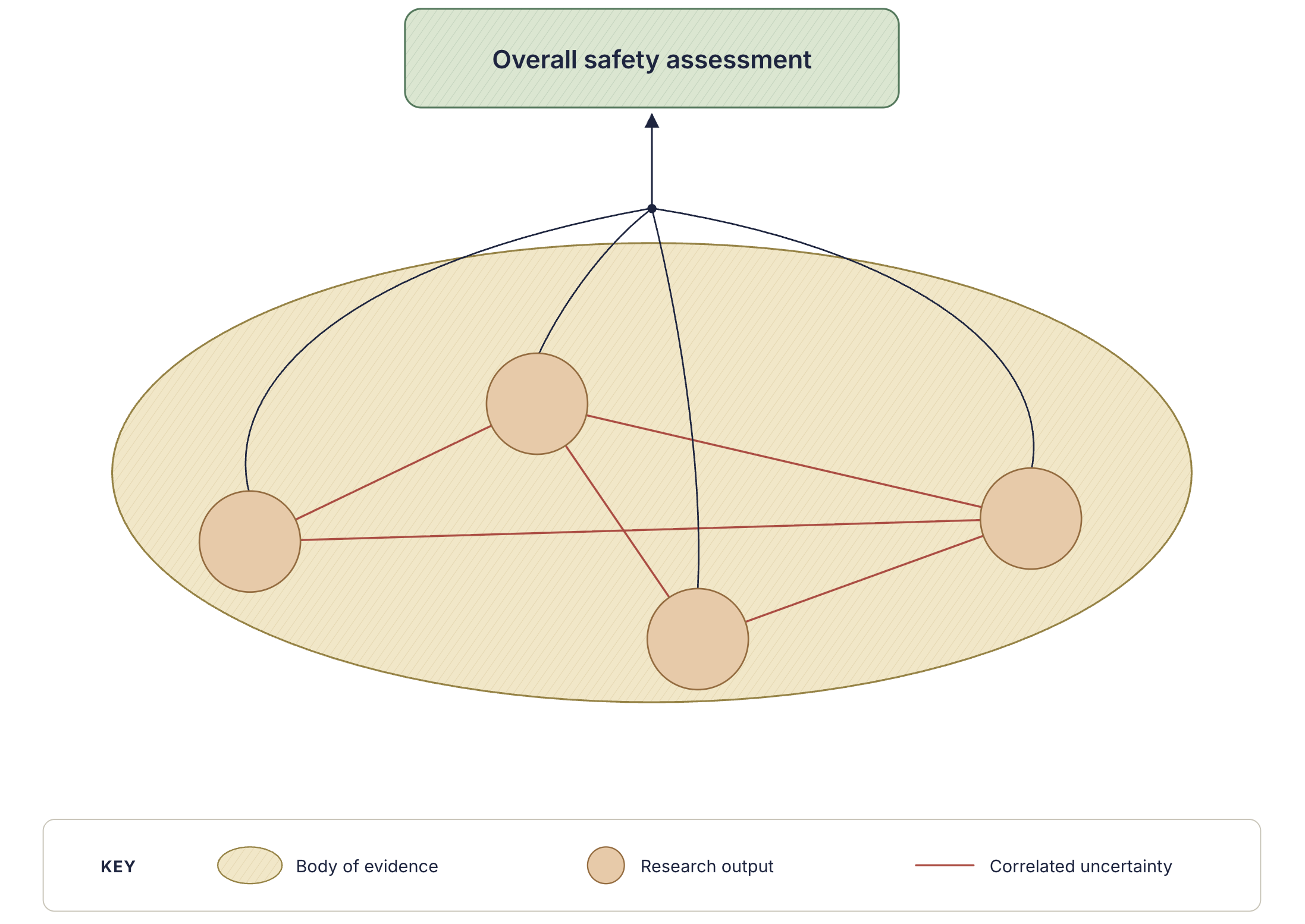}
  \caption{An overall safety assessment supported by four correlated research outputs (orange circles). The red lines indicate correlated uncertainty between research outputs. This correlated uncertainty makes combining judgements from each research output into an overall safety assessment difficult. Incorrectly assuming independence between research outputs produces an overconfident OSA which could lead to alignment failures.}
  \label{fig:4}
\end{subfigure}
\end{figure}

Current scalable oversight protocols assume that a single human can provide a clear and correct judgement for each subtask \citep{irving2018ai, christiano2018supervising, leike2018scalable}. In reality, for hard-to-supervise fuzzy tasks some subtasks are likely to remain difficult judgement problems where human experts disagree (Fig. \ref{fig:3}). Decomposition is still useful, however, as long as each subtask is easier for humans to correctly judge than the main task. The degree to which this holds for alignment-relevant hard-to-supervise fuzzy research tasks is an open question.

\hypertarget{correlated-uncertainty-makes-scalable-oversight-for-osas-harder}{%
\subsubsection{Correlated uncertainty makes scalable oversight for OSAs harder}\label{correlated-uncertainty-makes-scalable-oversight-for-osas-harder}}

Existing scalable oversight protocols do not satisfactorily deal with correlated uncertainty (Sec. \ref{aggregation-level-failures}) when recombining subtasks and are therefore not appropriate for evaluating OSAs. In most protocols, the recombination of subtasks is straightforward (debate, for instance, decomposes a claim into a deductive argument \citep{irving2018ai}). For OSAs, however, correctly combining subtasks is more difficult: The output of each subtask is a probability estimate (e.g. the probability that Agent N+1 responds honestly to queries of a certain type given a particular research output) and subtasks have a complicated correlation structure. A scalable oversight protocol for OSAs must therefore accurately model this correlation structure in order to correctly recombine subtasks (Fig. \ref{fig:4}).

\hypertarget{conclusion}{%
\section{Conclusion and future research}\label{conclusion}}

Even if agents are not scheming to sabotage alignment research they are likely to produce research that looks compelling but in fact contains undetected, systematic errors. These errors can occur in individual outputs or their aggregation into OSAs. We should expect these errors to be harder to identify than in human-generated alignment research. In this scenario, research could survive extensive human checks and strongly indicate that a next-generation model is safe to deploy but in fact be catastrophically wrong.

To prevent this, agents must reliably perform well on the hard-to-supervise fuzzy research tasks that are pervasive in alignment (such as making correct inferences about alignment from research outputs that target alignment proxies, or compiling OSAs from bodies of correlated evidence). This is because we cannot rely on iteration to resolve errors. The leading methods for training agents on hard-to-supervise fuzzy tasks have unresolved open problems. These are:

\textbf{Generalisation:} This technique relies on generalisation from easier training proxies to hard-to-supervise fuzzy tasks. Validating training proxies, however, requires generalisation behaviour to be predictable or performance on the hard-to-supervise fuzzy task to be evaluable.

\textbf{Scalable oversight:} This technique enables performance on hard-to-supervise fuzzy tasks to be evaluated by decomposing the tasks into easier-to-evaluate subtasks. It does not, however, completely remove the requirement for humans to make fuzzy judgement calls (although the hope is that remaining judgement calls will be easier). Existing protocols assume that recombining subtasks is straightforward. In reality, for OSAs, subtasks are likely to be highly correlated, making recombination challenging.

At the moment, we can only reliably test our alignment solutions on weaker-than-human models, where the alignment risk is low and we are able to judge the effectiveness of our techniques (misalignment in current models is often easy to catch as they often verbalise intent in chain of thought or take obviously deceptive actions). These techniques may not scale beyond human level and so we must find methods for evaluating the effectiveness of alignment solutions for superhuman models.

We see several avenues for further research, which we outline below. Most of these research projects require a proxy for a hard-to-supervise fuzzy task where the ground truth is known. Two options for proxies are:

\begin{enumerate}
\def\labelenumi{\arabic{enumi}.}
\item
  \textbf{Relative hard-to-supervise fuzzy tasks: }Tasks that are easy to supervise for a domain expert but hard to supervise for a non-expert. Expert evaluation provides a ground truth and we can test uplift of non-experts due to different scalable oversight protocols.
\item
  \textbf{Completed research results: }Completed research problems that are beyond an agent's knowledge cutoff contain examples of hard-to-supervise fuzzy tasks with known solutions.
\end{enumerate}

In addition to making progress on generalisation and scalable oversight, another possible research direction is to develop better ways of measuring agent performance on hard-to-supervise fuzzy tasks. Below is a non-comprehensive list of follow-up research ideas within each of these three areas. 

\textbf{Measuring agent capabilities}

\begin{enumerate}
\def\labelenumi{\arabic{enumi}.}
\item
  \textbf{Recreate completed research projects.} Take logs from completed research projects (intermediate documents, research meeting notes, code etc.) and measure agent ability to continue with the research project against the human baseline.
\item
  \textbf{Test agent prediction performance over datasets of correlated-events. }This directly tests whether agents are able to correctly combine correlated subtasks.
\item
  \textbf{Empirical studies of optimal human-agent team structure. }Use non-experts to solve completed research problems beyond the model's knowledge cutoff with agent assistance. Vary team structure and review methods. Measure the effectiveness of different structures at reproducing correct results.
\end{enumerate}

\textbf{Generalisation}

\begin{enumerate}
\def\labelenumi{\arabic{enumi}.}
\item
  \textbf{Simulated generalisation experiments} where different training proxies are tested using agent performance on completed research problems from beyond the knowledge cutoff of models. This could be done with weaker models with earlier knowledge cutoffs (increasing the number of available completed research results).
\item
  \textbf{Mechanistic understanding of generalisation} via whitebox methods, for example in mechanistic interpretability or learning theory.
\end{enumerate}

\textbf{Scalable oversight}

\begin{enumerate}
\def\labelenumi{\arabic{enumi}.}
\item
  \textbf{Compactification of research paper corpora.} An alternative to producing OSAs by aggregating results from a large body of research is to have an agent periodically produce a small number of completely new research outputs designed to capture the information that was learned and be maximally informative (a compactification of a research corpus). A compactification would be scored on a separate agent's ability to predict conclusions from the original research corpus using just the compactification. This metric could be optimised in training to produce better compactifications. This process would result in a small sample of maximally informative research outputs that human reviewers could more easily check and remain abreast of research developments. It also reduces the problem of correlated uncertainty by decreasing the number of research outputs required for an OSA.
\item
  \textbf{Develop and test new scalable oversight protocols} that deal with correlated uncertainty when recombining subtasks. Test uplift of non-experts on relative hard-to-supervise fuzzy tasks with complicated correlation structure and compare to existing protocols.
\item
  \textbf{Test different human scaffolds} for uplifting non-expert performance on relative hard-to-supervise fuzzy tasks.
\item
  \textbf{Red team automated alignment programs.} The red team prompts an agent to hide errors in a research paper corpus and the blue team attempts to catch these errors with agent assistance. This project would catalogue the type and severity of errors that evade detection and the effectiveness of different review processes.
\end{enumerate}

Successfully resolving the open problems we identify in this paper would increase the likelihood that automated alignment programs are able to provide effective and trustworthy alignment solutions.

\section*{Acknowledgments}
Charlie Griffin and Martin Soto contributed substantially to the core content of this paper. We are also grateful to David Africa, Joshua Clymer, Benjamin Hilton, Evan Hubinger, Daniel Kokotajlo, Jan Leike, Simon Marshall, and Konstantinos Voudouris  for valuable review and discussions.

\bibliography{references}


\end{document}